\pdfoutput=1

\documentclass[11pt]{article}

\usepackage[preprint]{acl}

\usepackage{times}
\usepackage{latexsym}

\usepackage[T1]{fontenc}

\usepackage[utf8]{inputenc}

\usepackage{microtype}

\usepackage{inconsolata}

\usepackage{graphicx}

\usepackage{hyperref}
\usepackage{url}
\usepackage{xcolor}

\usepackage{graphicx}
\usepackage{CJKutf8}
\usepackage{multirow}
\usepackage{array}
\usepackage{tabularx}
\usepackage{makecell}
\usepackage{amsmath} 
\usepackage{amssymb}
\usepackage{tablefootnote}
\usepackage{subcaption}
\usepackage{diagbox}
\usepackage{graphicx}
\usepackage{algorithm}
\usepackage[noend]{algpseudocode}
\usepackage{booktabs}
\usepackage{xcolor}
\usepackage{color, soul} 
\usepackage{colortbl}
\usepackage{makecell}
\definecolor{orange}{rgb}{1,0.5,0}

\newcolumntype{L}[1]{>{\raggedright\arraybackslash}p{#1}}
\newcolumntype{C}[1]{>{\centering\arraybackslash}p{#1}}
\newcolumntype{R}[1]{>{\raggedleft\arraybackslash}p{#1}}

\DeclareSymbolFont{extraup}{U}{zavm}{m}{n}
\DeclareMathSymbol{\varheart}{\mathalpha}{extraup}{86}
\DeclareMathSymbol{\vardiamond}{\mathalpha}{extraup}{87}

\usepackage{amsmath}
\usepackage{amssymb}
\usepackage{wasysym}

\DeclareSymbolFont{extraup}{U}{zavm}{m}{n}
\DeclareMathSymbol{\varspadesuit}{\mathalpha}{extraup}{83}
\DeclareMathSymbol{\varheartsuit}{\mathalpha}{extraup}{86}
\DeclareMathSymbol{\vardiamond}{\mathalpha}{extraup}{87}
\DeclareMathSymbol{\varclubsuit}{\mathalpha}{extraup}{88}

\definecolor{purple}{RGB}{128, 0, 128}

\title{Large Language Models Enhanced by Plug-and-Play Syntactic Knowledge for Aspect-based Sentiment Analysis}

\author{
    Yuanhe Tian$^{\varheart}$, \hspace{0.1cm}
    Xu Li$^{\spadesuit}$, \hspace{0.1cm}
    Wei Wang$^{\varclubsuit}$, \hspace{0.1cm}
    Guoqing Jin$^{\Diamond}$, \hspace{0.1cm}
    Pengsen Cheng$^{\Delta}$, \hspace{0.1cm}
    Yan Song$^{{\spadesuit}*}$
    \\
    $^{\varheart}$University of Washington 
    \hspace{0.1cm}
    $^{\spadesuit}$University of Science and Technology of China \\
    $^{\varclubsuit}$China Resources Digital Technology \hspace{0.1cm}
    $^{\Diamond}$People’s Daily Online \hspace{0.1cm}
    $^{\Delta}$Sichuan University \\
    $^{\varheart}$\texttt{yhtian@uw.edu} \hspace{0.1cm}
    $^{\spadesuit}$\texttt{lixu123@mail.ustc.edu.cn} \hspace{0.1cm}
    $^{\varclubsuit}$\texttt{ww.cs.tj@gmail.com} \\
    $^{\Diamond}$\texttt{jinguoqing@people.cn} \hspace{0.1cm}
    $^{\Delta}$\texttt{chengpengsen@scu.edu.cn} \hspace{0.1cm}
    $^{\spadesuit}$\texttt{clksong@gmail.com} 
}

\begin{document}
\maketitle

\renewcommand{\thefootnote}{\fnsymbol{footnote}}
\footnotetext[1]{Corresponding author.}
\renewcommand{\thefootnote}{\arabic{footnote}}

\begin{abstract}

Aspect-based sentiment analysis (ABSA) generally requires a deep understanding of the contextual information, including the words associated with the aspect terms and their syntactic dependencies.
Most existing studies employ advanced encoders (e.g., pre-trained models) to capture such context, especially large language models (LLMs).
However, training these encoders is resource-intensive, and in many cases, the available data is insufficient for necessary fine-tuning.
Therefore it is challenging for learning LLMs within such restricted environments and computation efficiency requirement.
As a result, it motivates the exploration of plug-and-play methods that adapt LLMs to ABSA with minimal effort.
In this paper, we propose an approach that integrates extendable components capable of incorporating various types of syntactic knowledge, such as constituent syntax, word dependencies, and combinatory categorial grammar (CCG).
Specifically, we propose a memory module that records syntactic information and is incorporated into LLMs to instruct the prediction of sentiment polarities. 
Importantly, this encoder acts as a versatile, detachable plugin that is trained independently of the LLM.
We conduct experiments on benchmark datasets, which show that our approach outperforms strong baselines and previous approaches, thus demonstrates its effectiveness.\footnote{The code and relevant resources are released at \url{https://github.com/synlp/LLM-Plugin-ABSA}.}

\end{abstract}

\section{Introduction}
\label{intro}

Aspect-based sentiment analysis (ABSA) is the task that predicts the sentiment of an aspect term, rather than the entire sentence, on the fine-grained level.
For example, the sentiment of the aspect terms ``\textit{price}'' and ``\textit{laptops}'' in the sentence ``\textit{The products are quite affordable, but the laptops very easily get hot.}'' are \textit{positive} and \textit{negative}, respectively, whereas the sentiment of the entire sentence tends to be \textit{negative}.
Therefore, this task is important in many real-world applications, such as analyzing the product review of users and monitoring the sentiment change of them on social media, which attracts much attention in recent years \cite{tang-etal-2020-dependency,chen-etal-2022-discrete,wang-etal-2023-reducing,luo2024panosent,tian2025representation}.

\begin{figure*}[t]
  \centering
  \includegraphics[width=1\textwidth, trim=0 10 0 0]{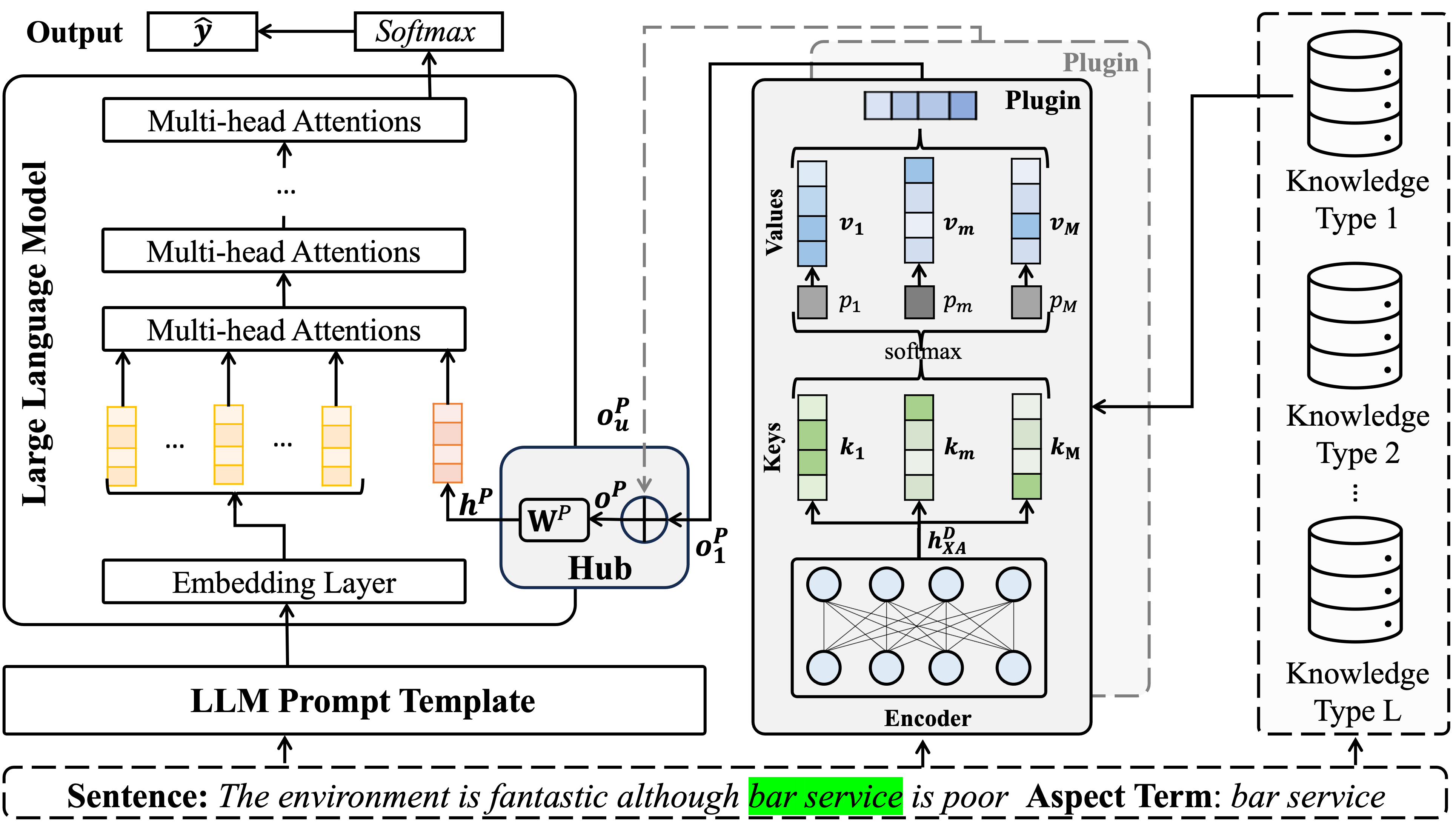}
  \caption{
  The overall architecture of the proposed approach of pluginning LLMs for ABSA.
  The left part illustrates the LLM and its prompt template for ABSA.
  The right part demonstrates the memory-based LLM plugin for different types of knowledge.
  The aspect term in the sentence is highlighted in green background.
  The output vector of the plugin is incorporated into the LLM through a hub module to integrate different types of knowledge.
  }
  \vskip -0.8em
  \label{fig:model}
\end{figure*}

For this task,
the key is 
to have a good modeling of contextual information.
Many previous studies use advanced encoders (e.g., Transformer \cite{transformer}) to capture contextual information and further incorporate external knowledge to help their models understand the aspect terms along with their context \cite{wang2016attention,ma2018targeted,liang-etal-2019-context,tang-etal-2020-dependency,liang-etal-2021-iterative-multi,zhang-etal-2022-ssegcn,ma-etal-2023-amr}.
Although
recent success of large language models (LLMs) on natural language processing (NLP) \cite{ouyang2022training,touvron2023llama-2,taori2023alpaca} provides much stronger context encoding power for ABSA,
they either require heavy training or appropriate
design of prompts to instruct the LLMs to perform the task \cite{li2024iterative,zheng2024instruction},
which is resource-intensive in terms of both computation and human involvement.
There are studies \cite{zhao2021knowledge,brauwers2022survey,dekker2023knowledge} that leverage knowledge, such as frequent aspect term and sentiment pairs, to improve ABSA, whereas these types of knowledge may not perform well with LLMs since these types of knowledge are fixed and lack the flexibility to adapt to the rich, dynamic contexts captured by LLMs.
From other perspectives,
many ABSA studies also show that syntactic information is helpful for ABSA \cite{huang2019syntax,zhang2019aspect,wang-etal-2020-relational,tang-etal-2020-dependency},
for the reason that syntax-driven word associations provide useful context information that plain encoders usually cannot do.
Therefore, it is straightforward to
explore lightweight and versatile methods to train and adapt LLMs with syntactic knowledge for ABSA,
so that both the power of context modeling from LLMs and formal linguistic knowledge are expected to be leveraged with a plug-and-play design.

In this paper, we propose to pluginning LLMs for ABSA, a memory-based plugin is utilized to explicitly incorporate different types of knowledge,
which is opted to be trained independently without LLMs.
In detail, the plugin takes the input text as well as the representation of a particular type of knowledge, then produces a vectorized output that encodes the corresponding syntactic knowledge.
A hub module is proposed to integrate the knowledge vector into the LLM decoding process, so as to instruct the LLM to predict the sentiment polarity
with the help of the plugin.
Note that,
when there are multiple types of knowledge, one can use several copies of the plugin to model each knowledge input individually, and the outputs from different plugins are fused in the hub module and then sent to LLMs.
We validate the effectiveness of the proposed plugin with three types of syntactic information, namely, dependency relations, constituent syntax, and CCG supertags, all of which have proven effective in capturing rich contextual evidence for ABSA \cite{tang-etal-2020-dependency,wang-etal-2020-relational,tian2021aspect}.
Experiment results and further analyses on benchmark datasets for ABSA demonstrate the effectiveness of the proposed approach, where our approach outperforms strong baselines and achieves state-of-the-art on them.

\section{The Proposed Approach}
\label{approach}

Our approach for ABSA utilizes a memory-based plugin to enhance LLMs for ABSA, where the plugin is able to encode different types of knowledge and the encoded vector is incorporated into the LLM through a specially designed hub module.
Figure \ref{fig:model} presents the overall architecture of the proposed approach, where the given aspect term in the example input sentence is highlighted in green background.
The left side of the figure illustrates the LLM for ABSA.
The right side illustrates the memory-based plugin to encode different types of knowledge.
Therefore, our approach for ABSA is formalized as
\begin{equation}
\setlength\abovedisplayskip{4pt}
\setlength\belowdisplayskip{4pt}
    \widehat{y} = \underset{y \in \mathcal{T}}{\arg\max} \ 
    p(y | \mathcal{X}, \mathcal{A}, \mathcal{H}(\mathcal{P}(\mathcal{X}, \mathcal{A}, \mathbf{K}, \mathbf{V})))
\end{equation}
when the plugin $\mathcal{P}$ is used with an LLM 
$p$ by predicting a particular sentiment polarity $y \in \mathcal{T}$ given $\mathcal{X}$ and $\mathcal{A}$;
$\mathcal{H}$ denotes the hub module,
$\mathbf{K}$ and $\mathbf{V}$ are key and value vectors that carry knowledge information and are processed by the plugin.

In the following subsections, we firstly illustrate the memory-based plugin, then describe the hub on how it incorporates the plugin output into the LLM,
and finally present the LLM decoding process when it is integrated with the plugin and hub.

\subsection{The Memory-based Plugin}
\label{sec: plugin}

The memory-based plugin contains an encoder $\mathcal{E}$ with a memory component to encode the knowledge.
It takes the input text (which includes $\mathcal{X}$ and $\mathcal{A}$) and a list of key vectors $\mathbf{K}=[\mathbf{k}_1, \cdots \mathbf{k}_m, \cdots \mathbf{k}_M]$, and a list of value vectors $\mathbf{V}=[\mathbf{v}_1, \cdots \mathbf{v}_m, \cdots \mathbf{v}_M]$, where the number of keys and values (i.e., $M$) are the same, and produce a vector that encodes the knowledge.
\textcolor{black}{
Specifically, the plugin uses $\mathcal{E}$ to encode the input text (which includes $\mathcal{X}$ and $\mathcal{A}$) and leverages keys and values to represent different context information instances (e.g., the word dependencies or the latent topics).
During the modeling process, the encoded input representation is used as the query vector to weigh different context information instances.
}

\textcolor{black}{
Specifically, we firstly concatenate the input sentence $\mathcal{X}$ and the aspect term $\mathcal{A}$ and use the encoder $\mathcal{E}$ (e.g., BERT \cite{devlin2019bert}) to compute the vector representation $\mathbf{h}_{XA}$ of them by
\begin{equation} \label{eq:r}
\setlength\abovedisplayskip{5pt}
\setlength\belowdisplayskip{5pt}
\mathbf{h}_{XA} = \mathcal{E}(\mathcal{X} \oplus \mathcal{A})
\end{equation}
where $\oplus$ denotes the concatenation operation\footnote{For example, if the encoder is BERT, the input of BERT is ``\textit{[CLS] $\mathcal{X}$ [SEP] $\mathcal{A}$ [SEP]}'' a special word ``\textit{[SEP]}'', where the special words (i.e., ``\textit{[CLS]}'' and ``\textit{[SEP]}'') are added to mark the boundary of the input sentence and the aspect term $\mathcal{A}$. Meanwhile, the $\mathbf{h}_{XA}$ is the hidden vector for ``\textit{[CLS]}'' obtained from the last layer of BERT.}.
Then, we compute the weight $p_m$ for the $m$-th value vector (which corresponds to the $m$-th context information instance) by
\begin{equation} \label{eq: p_ij}
\setlength\abovedisplayskip{5pt}
\setlength\belowdisplayskip{5pt}
    p_{m} = \frac{exp(\mathbf{h}_{XA} \cdot {\mathbf{k}_{m}})}{\sum_{m=1}^{M} exp(\mathbf{h}_{XA} \cdot {\mathbf{k}_{m}})}
\end{equation}
Afterwards, we compute the weighted sum of the value embeddings and obtain the output $\mathbf{o}^P$ of the plugin through
\begin{equation} \label{eq:r}
\setlength\abovedisplayskip{5pt}
\setlength\belowdisplayskip{5pt}
\mathbf{o}^P = \sum_{m=1}^{M} p_{m} \cdot {\mathbf{v}_{m}}
\end{equation}
Finally, $\mathbf{o}^P$ is regarded as the output of the plugin and is fed into the hub module to incorporate $\mathbf{o}^P$ into the LLM.
}
It is worth noting that our plugin is able to process different types of knowledge when they are processed in the same way.
Therefore, when there are multiple knowledge sources, we use several plugins by feeding each one with a specific knowledge type following the aforementioned process.
%
The output of the $u$-th plugin is denoted as $\mathbf{o}^{P}_u$, where $u \in [1, U]$ with $U$ denoting the total number of plugins.

\subsection{The Hub}

To incorporate a plugin with LLMs,
we use a hub module to connect them,
which is designed in a straightforward manner with
an optional vector concatenation operation and a multi-layer perceptron (MLP).
Once the output of the plugin $\mathbf{o}^P$ is obtained, the hub passes $\mathbf{o}^P$ through the MLP to compute the hidden vector $\mathbf{h}^P$ through
\begin{equation} \label{eq: h_p}
\setlength\abovedisplayskip{5pt}
\setlength\belowdisplayskip{5pt}
    \mathbf{h}^{P} = MLP (\mathbf{o}^P)
\end{equation}
where $\mathbf{h}^{P}$ is used in the LLM decoding process to instruct the LLM to perform ABSA task.
For the scenarios where multiple plugins are used
(i.e., $\mathbf{o}^{P}_1 \cdots \mathbf{o}^{P}_U$ are obtained),
the hub concatenates their outputs by $\mathbf{o}^{P} = \oplus_{u=1}^{U} \mathbf{o}^{P}_u$,
and follow the same process in Eq. (\ref{eq: h_p}) to compute $\mathbf{h}^{P}$.

\subsection{Pluginning LLMs for ABSA}
\label{sec: train}

There are two strategies to plugin LLMs: the first one is that we tune the plugin together with the LLM on the task data;
the second one is that we train the plugin independently and then plug it into the LLM without further tuning.
The following text illustrates the details of the two strategies.

\textcolor{black}{
For the first strategy, we firstly use a prompt template $\mathcal{T}$ and fill it with the input sentence $\mathcal{X}$ and the aspect term $\mathcal{A}$.
Next, we use the embedding layer of the LLM to convert all words in the filled prompt template into their embeddings, which results in an embedding matrix $\mathbf{E}$.
Then, we concatenate the embedding matrix $\mathbf{E}$ with the vector $\mathbf{h}^P$ obtained from the hub through Eq. (\ref{eq: h_p}) and get the new embedding matrix $\mathbf{E}'=[\mathbf{E}, \mathbf{h}^P]$.
Afterwards, we feed $\mathbf{E}'$ into the first multi-head attention layer of the LLM and follow the standard LLM decoding process to compute the output hidden vector from the last layer.
Finally, we pass the hidden vector through a softmax classifier to predict the sentiment polarity $\widehat{y}$.
We compare the predicted label $\widehat{y}$ with the gold standard label $y^*$, compute the loss, and optimize the plugin and the hub accordingly, where the parameters in the LLM are fixed.
}

\textcolor{black}{
For the second strategy, we directly train the plugin on ABSA and use the output of the plugin to instruct the LLM to predict the ABSA label in inference.
Specifically, to train the plugin on ABSA, we compute the sum of $\mathbf{o}^P$ and the query vector $\mathbf{h}_{XA}$ and pass the resulting vectors through a softmax classifier $f_{P}$ to predict the sentiment $\widehat{y}^{P}$ of the input, which is formulated as
\begin{equation} \label{eq: y^p}
\setlength\abovedisplayskip{5pt}
\setlength\belowdisplayskip{5pt}
    \widehat{y}^{P} = f_{P} (\mathbf{o}^P \oplus \mathbf{h}_{XA})
\end{equation}
We compute the loss by comparing the prediction $\widehat{y}^{P}$ with the gold standard $y^*$ and update the parameters in the plugin accordingly.
In inference, when the plugin is working with the LLM, we firstly use the plugin to predict the sentiment label $\widehat{y}^{P}$.
Then we use the input sentence $\mathcal{X}$, the aspect term $\mathcal{A}$, and the plugin prediction $\widehat{y}^{P}$ to fill a prompt template $\mathcal{T}'$ that has a particular slot to fill the plugin prediction\footnote{For example, the template could contain a sentence saying ``\textit{The prediction of the plugin is [the plugin prediction]}'' (the ``\textit{[the plugin prediction]}'' will be replaced by $\widehat{y}^{P}$).}.
If there are multiple plugins, the template will have a slot for each one to present its prediction.
Afterwards, we feed the filled template into the LLM and follow the standard decoding process to get the prediction $\widehat{y}$.}

\textcolor{black}{
To summarize, the selection of training strategy mainly depends on computational resources and practical application scenarios.
Strategy 1 is suitable when sufficient computational resources and fine-tuning capabilities are available, potentially leading to slightly better performance. 
Strategy 2 is designed for scenarios with limited computational resources, allowing rapid adaptation without heavy computation. 
}

\begin{table}[t]
\begin{center}
        \centering
        \begin{tabular}{l l | r | r| r}
        \toprule
        \multicolumn{2}{c|}{\textbf{Dataset}} & \multicolumn{1}{c|}{\textbf{Pos. \#}} & \multicolumn{1}{c|}{\textbf{Neu. \#}} & \multicolumn{1}{c}{\textbf{Neg. \#}} \\
        \midrule
        \multirow{2}{*}{\textbf{LAP14}} & Train & 994 & 464 & 870 \\
        & Test & 341 & 169 & 128 \\
        \midrule
        \multirow{2}{*}{\textbf{REST14}} & Train &2,164 &637 &807 \\
        & Test &728 &196 &182 \\
        \midrule
        \multirow{2}{*}{\textbf{REST15}} & Train &907 &36 &254 \\
        & Test &326 &34 &207 \\
        \midrule
        \multirow{2}{*}{\textbf{REST16}} & Train &1,229 &69 &437 \\
        & Test &469 &30 &114 \\
        \midrule
        \multirow{3}{*}{\textbf{MAMS}} & Train & 3,380 & 5,042 & 2,764 \\
        & Dev & 403 & 604 & 325 \\
        & Test & 400 & 607 & 329 \\
        \midrule
        \midrule
        \multirow{2}{*}{\textbf{SE-15}} &
        Train & 2,301 & 259 & 1,168 \\
        & Test & 1,238 & 136 & 759 \\
        \bottomrule
        \end{tabular}
\end{center}
\vskip -0.3cm
\caption{\label{tab: dataset}
        The statistics of the datasets, where the number of instances with different sentiment polarities in the training, development, and test sets are reported.
        }
 \vskip -1em       
\end{table}

\section{Experiment Settings}

\subsection{Datasets}

Following previous studies, we run different models on five English benchmark datasets from different domains for ABSA, i.e., 
LAP14 and REST14 \cite{pontiki-etal-2014-semeval}, 
REST15 \cite{pontiki2015semeval}, 
REST16 \cite{pontiki2016semeval}, 
and MAMS\footnote{We use the ATSA part of MAMS obtained from \url{https://github.com/siat-nlp/MAMS-for-ABSA}.} \cite{jiang-etal-2019-challenge}.
Specifically, LAP14 contains laptop computer reviews, REST14, REST15, REST16, and MAMS is collected from online reviews of restaurants.
\textcolor{black}{
In addition, to evaluate the generalization of our approach in cases where the aspect term is not given, we also run experiments on SemEval 2015 Task 12 (SE-15) \cite{pontiki2015semeval} for joint ABSA, where the model needs to predict both the aspect term and the sentiment polarity.}
We use their official train/dev/test splits\footnote{Since LAP14, REST14, REST115, and REST16 do not have their official development sets, we randomly sample 10\% of the training sets to be the development sets.} for all datasets.
We report the statistics (i.e., the numbers of aspect terms with ``\textit{positive}'', ``\textit{negative}'', and ``\textit{neutral}'' sentiment polarities) of the five datasets in Table \ref{tab: dataset}.

\begin{figure}[t]
 \centering
 \includegraphics[width=0.45\textwidth, trim=0 20 0 10]{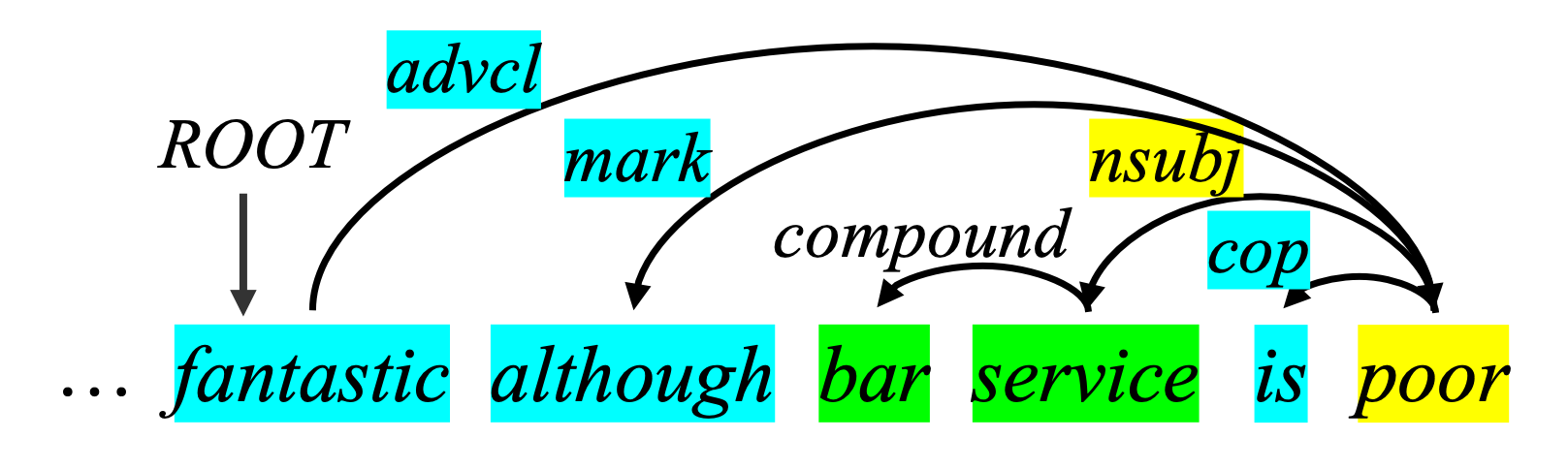}
  \caption{
  \textcolor{black}{
  A part of the dependency tree for the sentence ``\textit{The environment is fantastic although bar service is poor.}''.
  The aspect term is highlighted by the green background color.
  The first- and second-order dependencies are marked by the yellow and blue background colors, respectively.
  }
  }
  \vskip -1.2em
  \label{fig: dep-example}
\end{figure}

\subsection{Plugin with Constituent Syntax}
\label{sec: plugin_constituent}

\textcolor{black}{
To capture constituent-based syntactic information, we first obtain the constituency parse tree of the input sentence using an off-the-shelf parser (e.g., the constituency parser of Stanza \cite{qi-etal-2020-stanza}\footnote{\url{https://github.com/stanfordnlp/stanza}.}). 
We extract the longest phrases that contain the aspect term with a word-based length of less than 10.
We denote the words in the phrase as $c_1 \cdots c_{M_c}$ and the syntactic label of the phrase as $c_s$.
We regard the words $c_1 \cdots c_{M_c}$ as the keys and the combination of the words and the label $c_s$ as the values, and utilize two trainable embedding matrices to map them to the key embeddings and value embeddings, respectively (the $m_c$-th key and value vectors are $\mathbf{k}^{C}_{m_c}$ and $\mathbf{v}^{C}_{m_c}$, respectively).
We use Eq. (\ref{eq: p_ij}) and Eq. (\ref{eq:r}) to compute the output of the plugin, which is denoted as $\mathbf{o}^{C}$.
}

\subsection{Plugin with Word Dependencies}
\label{knowledge-ext}

Similar to existing studies that leverage word dependencies for ABSA, we employ an off-the-shelf toolkit, e.g., Stanza \cite{qi-etal-2020-stanza}\footnote{\url{https://github.com/stanfordnlp/stanza}.}, to generate dependency trees for input texts.
Once the parse tree of an input sentence is obtained, each word would have exactly one head word\footnote{For the root word of the dependency tree, we create a special word ``\textit{ROOT}'' and use it as its head word.} and may have one or more dependent words\footnote{For the leaf words, they do not have dependent words.}, and there is exactly one dependency path connecting each pair of words in the sentence.
\textcolor{black}{
In our experiments, we consider first- and second-order dependencies\footnote{We also try first-order dependencies only or extend it to third-order dependencies. They show worse performance than the first- and second-order setting.} that are relevant to the aspect term.
Specifically, for the first-order dependencies, we extract all words $w$ that have a direct dependency connection with any word $a_i$ in the aspect term from the dependency parse tree.
We also extract the relation type $r$ between $a_i$ and $w$, and pair the context word $w$ with the relation type $r$ to form $(w, r)$ ($w$ is either the head or the dependent word of $a_i$).
For second-order dependencies, we extract the words $w$ where there is an intermediate word $w'$ such that there is a dependency path ``$a_i$ -- $w'$ -- $w$'' between $a_i$ and $w$.
We pair the word $w$ with the dependency relation type $r$ between $w'$ and $w$ and construct a pair $(w, r)$ for $a_i$.
Next, we collect all extracted pairs of all words in the aspect term and filter out the ones where the context word $w$ belongs to the aspect term.
This process allows us to filter out the contextual information within the aspect term and thus focus more in other contextual information.
Using the sentence and dependency tree in Figure \ref{fig: dep-example} as an example, the extracted dependencies for the aspect term ``\textit{bar service}'' are \textit{(poor, nsubj)}, \textit{(is, cop)}, \textit{(although, mark)}, and \textit{(fantastic, advcl)}.
Then, we collect the remaining pairs and rank them based on their frequency in the training data.
We select the most frequent $M_d$ pairs and denote the pair list as $(w_1, r_1), \cdots (w_{m_d}, r_{m_d}) \cdots (w_{M_d}, r_{M_d})$, where $(w_{m_d}, r_{m_d})$ is the $m_d$-th pair.
}

\textcolor{black}{
To model the dependencies through the plugin, for each pair $(w_{m_d}, r_{m_d})$, we 
use two trainable embedding matrices to separately map $w_{m_d}$ and the combination\footnote{For example, for the pair \textit{(poor, nsubj)}, the combination of them is \textit{poor-nsubj}.} of $w_{m_d}$ and $r_{m_d}$ to their embeddings, which are denoted as $\mathbf{k}^{D}_{m_d}$ and $\mathbf{v}^{D}_{m_d}$.
We regard all $\mathbf{k}^D_{m_d}$ and $\mathbf{v}^{D}_{m_d}$ as the key vectors and value vectors in the plugin and follow Eq. (\ref{eq: p_ij}) and Eq. (\ref{eq:r}) to compute the output of the plugin, which is denoted as $\mathbf{o}^{D}$
Through this process, different dependency connections and relations are weighed, which allows the model to identify the important ones for ABSA and leverage them accordingly.
}

\subsection{Plugin with CCG Supertag}
\label{sec: plugin_ccg}

\textcolor{black}{
CCG supertags provide fine-grained syntactic information that goes beyond conventional POS tags.
For each token in the input sentence, we extract its CCG supertag via an off-the-shelf toolkit named NeST-CCG \cite{tian-etal-2020-supertagging}\footnote{\url{https://github.com/cuhksz-nlp/NeST-CCG}} and consider a context window of $\pm 3$ words around the aspect term. 
By pairing each word with its corresponding CCG supertag (e.g., ``bar\_N/N''), we generate a set of context features that capture both lexical and syntactic details. 
We regard each token and the combination of the token and the CCG supertags as keys and values, and then map them into their embeddings using dedicated trainable layers. 
Similarly, we utilize the plugin to model the information and obtain an aggregated representation \(\mathbf{o}^{S}\).
}

\begin{table*}[t]
\begin{center}
\centering
\scalebox{0.95}{
    \begin{tabular}{l | c c | c c |c c | c c | c c }
        \toprule
        \multirow{2}{*}{} & \multicolumn{2}{c|}{\textbf{LAP14}} & \multicolumn{2}{c|}{\textbf{REST14}} & \multicolumn{2}{c|}{\textbf{REST15}}  & \multicolumn{2}{c|}{\textbf{REST16}}  & \multicolumn{2}{c}{\textbf{MAMS}} \\ 
        \cline{2-11}
        \addlinespace[0.15cm]
        & ACC & F1 & ACC & F1 & ACC & F1 & ACC & F1 & ACC & F1  \\
        \midrule
        Qwen-2.5  
        & 82.76 & 78.59 
        & 85.71 & 76.90 
        & 86.06 & 73.92 
        & 91.24 & 82.49 
        & 81.32 & 81.70 \\

        \midrule

        \quad +P(C) 
        & 82.93 & 79.59 
        & 87.32 & 82.54 
        & 86.82 & 74.47 
        & 92.11 & 83.66 
        & 84.74 & 85.33  \\

        \quad +P(D) 
        & 83.10 & 79.73 
        & 87.65 & 82.82 
        & 87.45 & 74.41 
        & 92.08 & 83.82 
        & 84.86 & 85.21  \\

        \quad +P(S) 
        & 83.22 & 79.80 
        & 87.63 & 82.71 
        & 87.43 & 74.35 
        & 92.16 & 83.76 
        & 84.96 & 85.48  \\
        
        \midrule
        \quad +P(C) +P(D) +P(S)  & \textbf{83.33} & \textbf{79.95} & \textbf{87.98} & \textbf{82.91} & \textbf{87.83} & \textbf{75.52} & \textbf{93.27} & \textbf{84.27} & \textbf{85.93} & \textbf{86.37} \\  

        \midrule
        \midrule

        Qwen-2.5 (LoRA)  
        & 82.59 & 78.48 
        & 85.67 & 76.82 
        & 85.89 & 73.77 
        & 91.10 & 82.40 
        & 81.26 & 81.66 \\
        \midrule
        \quad +P(C) 
        & 82.90 & 79.52 
        & 87.28 & 82.45 
        & 86.76 & 74.49 
        & 92.01 & 83.59 
        & 84.61 & 85.27  \\
        \quad +P(D) 
        & 82.99 & 79.61 
        & 87.40 & 74.35 
        & 92.02 & 83.74 
        & 91.96 & 83.79 
        & 84.80 & 85.16  \\
        \quad +P(S) 
        & 83.13 & 79.75 
        & 87.57 & 82.63 
        & 87.37 & 74.30 
        & 92.11 & 83.78 
        & 84.89 & 85.53  \\
        \midrule
        \quad +P(C)+P(D)+P(S)  
        & \textbf{83.27} & \textbf{79.91} 
        & \textbf{87.91} & \textbf{82.89} 
        & \textbf{87.79} & \textbf{75.47} 
        & \textbf{93.21} & \textbf{84.22} 
        & \textbf{85.84} & \textbf{86.32} \\   
        
        \midrule
        \midrule
        
        LLaMA-2-7B  
        & 82.64 & 78.42 
        & 85.61 & 76.83 
        & 85.98 & 73.87 
        & 91.21 & 82.39 
        & 81.42 & 81.78 \\

        \midrule

        \quad +P(C) 
        & 82.97 & 79.64 
        & 87.30 & 82.50 
        & 86.85 & 74.44 
        & 92.03 & 83.57 
        & 84.69 & 85.27  \\

        \quad +P(D) 
        & 83.20 & 79.83 
        & 87.65 & 82.77 
        & 87.32 & 74.39 
        & 92.03 & 83.71 
        & 84.90 & 85.32  \\

        \quad +P(S) 
        & 83.29 & 79.85 
        & 87.68 & 82.80 
        & 87.41 & 74.36 
        & 92.12 & 83.68 
        & 85.02 & 85.39  \\

        \midrule

        \quad +P(C)+P(D)+P(S)  
        & \textbf{83.35} & \textbf{80.01} 
        & \textbf{87.92} & \textbf{82.88} 
        & \textbf{87.85} & \textbf{75.44} 
        & \textbf{93.32} & \textbf{84.29} 
        & \textbf{86.05} & \textbf{86.44} \\  
        
        \bottomrule
    \end{tabular}
}
\end{center}
\vspace{-0.3cm}
\caption{
\textcolor{black}{
Experiment results (i.e., the average accuracy and F1 scores) of different models on the test set of all datasets for ABSA, where different combinations of the constituent syntax (C), dependency knowledge (D), or the CCG supertags (S) are used.
The configurations with ``P($\cdot$)'' indicate that the plugin is used to model the knowledge.
}
}
\label{tab: overall}
\vskip -0.8em
\end{table*}


\subsection{Implementation Details}

\textcolor{black}{
Since a good text representation is able to enhance model performance on downstream tasks \cite{mikolov2013efficient,song2017learning,song2018joint,peters-etal-2018-deep,song2018complementary,devlin2019bert}, we employ pre-trained models for our experiments.
Specifically, we employ Qwen2.5 (1.5B) \cite{yang2024qwen2} and LLaMA-2 (7B) \cite{touvron2023llama-2} as the LLM in the experiments, and utilize BERT \cite{devlin2019bert} as the encoder of the plugin.\footnote{We obtain Qwen2.5, LLaMA-2, and BERT from 
\url{https://huggingface.co/Qwen/Qwen2.5-1.5B-Instruct}, \url{https://huggingface.co/meta-llama/Llama-2-7b-chat-hf}, and \url{ https://huggingface.co/google-bert/bert-base-uncased} respectively.}
We follow the default setting of these models in all experiments.
Specifically, for the BERT model, we use 12 layers of self-attention with 768-dimensional hidden vectors; 
for Qwen2.5, we use 28 layers of self-attention with 1536-dimensional hidden vectors;
for LLaMA-2, we use 32 layers of self-attention with 4096-dimensional hidden vectors.
}
We set the default number of keys and values (i.e., memory size) in the plugin as five (i.e., $M_d=5$ and $M_t=5$).
In addition, the loss function in our models is cross-entropy, and evaluation metrics are accuracy and F1 scores overall sentiment polarities, following the conventions widely used by previous studies \cite{tang-etal-2016-effective,he-etal-2018-effective,tian2021aspect}.
\textcolor{black}{
For each model, we fine-tune it with different random seeds three times and report the average performance on the test sets.
}

\section{Results and Analysis}

\subsection{Overall Results}
\label{sec: kvmn}

\textcolor{black}{
To explore the effect of our approach to pluginning LLMs with incorporating different types of knowledge, namely, the constituent syntax, word dependencies, and CCG supertags, we run models with various approaches to leverage different combinations of knowledge types.
The results (i.e., the accuracy and F1 scores) of different LLMs (i.e., Qwen-2.5 and LLaMA-2 with full parameter fine-tuning and Qwen-2.5 fine-tuned with LoRA) on the test set of all benchmark datasets are reported in Table \ref{tab: overall}.
For each LLM, there are five models.
The first model is the vanilla LLM that does not utilize any type of knowledge.
The second, third, and fourth models utilize a single plugin to leverage one of the constituent syntax (C), word dependencies (D), and CCG supertags (S), which are represented by ``+P(C)'', ``+P(D)'', and ``+P(S)'', respectively.
The last model uses three plugins to model all types of knowledge.
All models with the plugin are tuned following the first approach illustrated in Section \ref{sec: train} (i.e., the hub and the plugin are tuned on the training data with the parameters in the LLM fixed).
}

\begin{table*}[t]
\begin{center}
\centering
\scalebox{0.9}{
    \begin{tabular}{l|cc|cc|cc|cc|cc}
        \toprule
        \multirow{2}{*}{\textbf{Models}} & \multicolumn{2}{c|}{\textbf{LAP14}} & \multicolumn{2}{c|}{\textbf{REST14}} & \multicolumn{2}{c|}{\textbf{REST15}}  & \multicolumn{2}{c|}{\textbf{REST16}}  & \multicolumn{2}{c}{\textbf{MAMS}}            \\
        & ACC   & F1    & ACC   & F1    & ACC   & F1    & ACC   & F1    & ACC   & F1        \\
        \midrule
        \citet{liang-etal-2022-bisyn}
        & 82.91 & 79.38  
        & 87.94 & 82.43 
        & - & -
        & - & -
        & 85.85 & 85.49 
        \\

        \citet{tang-etal-2022-affective}
        & 81.83 & 78.26  
        & 87.31 & 82.37 
        & - & -
        & - & -
        & - & - 
        \\

        \citet{cao-etal-2022-aspect}
        & 82.75 & 79.95  
        & 87.67 & 82.59 
        & - & -
        & - & -
        & - & - 
        \\

        \citet{zhang-etal-2022-ssegcn}
        & 81.01 & 77.96  
        & 87.31 & 81.09 
        & - & -
        & - & -
        & - & - 
        \\
        
        \citet{chen-etal-2022-discrete}
        & 81.03 & 78.10  
        & 86.16 & 80.49 
        & 85.24 & 72.74
        & 93.18 & 82.32
        & - & - 
        \\
        \citet{ma-etal-2023-amr}
        & 81.96 & 79.10  
        & 87.76 & 82.44 
        & - & -
        & - & -
        & 85.59 & 85.06 
        \\
        \citet{zhang-etal-2023-span}
        & 81.80 & 78.46  
        & 87.09 & 81.15 
        & - & -
        & - & -
        & - & - 
        \\
        \citet{zhang-etal-2023-empirical}
        & - & 78.68  
        & - & 81.59 
        & - & -
        & - & -
        & - & 83.65 
        \\
        \citet{chai-etal-2023-aspect}
        & 82.12 & 78.82  
        & 87.86 & 82.41 
        & 86.74 & 75.05
        & \textbf{93.42} & 83.80
        & 85.10 & 84.65 
        \\ 
        \citet{wang-etal-2023-reducing}
        & 81.56 & 75.92  
        & 86.37 & 80.63 
        & 83.98 & 70.86
        & 91.45 & 78.12
        & 84.68 & 84.23 
        \\

        *\citet{li2024iterative}
        & 82.49 & 79.62  
        & 87.50 & 81.68 
        & 87.13 & 75.17
        & 92.95 & 82.83
        & - & - 
        \\

        *\citet{zhou2024comprehensive}
        & 81.04 & -  
        & 87.35 & - 
        & 87.27 & -
        & - & -
        & - & - 
        \\

        *\citet{li2025exploring}
        & 77.20 & -  
        & - & - 
        & - & -
        & - & -
        & - & - 
        \\
        \midrule

        RAG (LLaMA-2+C+D+S)
        & 78.40 & 71.30 
        & 86.28 & 71.09 
        & 84.31 & 70.81 
        & 89.79 & 73.83 
        & 81.18 & 82.08 
        \\

        \midrule
         Ours (LLaMA-2) & \textbf{83.35} & \textbf{80.01} 
        & \textbf{87.92} & \textbf{82.88} 
        & \textbf{87.85} & \textbf{75.44} 
        & 93.32 & \textbf{84.29} 
        & \textbf{86.05} & \textbf{86.44} \\
        \bottomrule
    \end{tabular}
}
\end{center}
\vspace{-0.3cm}
\caption{\label{tab: sota}
The comparison of our best model (i.e., the one with \textcolor{black}{LLaMA-2} and plugins to leverage all three types of syntactic knowledge with previous studies on all datasets.
The approach that uses LLMs is marked by ``*''.
\textcolor{black}{
The ``RAG'' approach means the extracted syntactic knowledge is represented in text and directly combined with the input to prompt the LLM to predict the sentiment polarities.
}
}
    \vskip -1em
\end{table*}

\begin{table}[t]
\begin{center}
\centering
    \begin{tabular}{l|c}
        \toprule
        & ACC           \\
        \midrule
        \citet{wallaart2019hybrid}
        & 81.7
        \\
        \citet{dekker2023knowledge}
        & 81.9
        \\
        \midrule
        Qwen-2.5 & 
        80.6 \\
        Ours (Qwen-2.5) & \textbf{82.3} \\
        \bottomrule
    \end{tabular}
\end{center}
\vspace{-0.3cm}
\caption{\label{tab: sota-joint}
\textcolor{black}{
Comparison of accuracy between our approach and existing studies on joint ABSA on the test set of SemEval 2015, where the result of our runs of the vanilla Qwen-2.5 is also reported for reference.
}
}
\vskip -1em
\end{table}

\textcolor{black}{
The following are some observations.
First, the performance consistently improves over the vanilla LLM baseline on all datasets if one of ``+P(C)'', ``+P(D)'' ``+P(S)'' is used, which presents the effectiveness of using the plugin to model one types of contextual information for ABSA.
Second, our full model with all three plugins achieves better performance than the settings where only one plugin is used.
This shows that the plugin is able to coordinate well with each other to incorporate different types of knowledge.
}

\begin{table*}[t]
\begin{center}
\centering
\scalebox{0.93}{
    \begin{tabular}{l | c c | c c |c c | c c | c c | c}
        \toprule
        \multirow{2}{*}{} & \multicolumn{2}{c|}{\textbf{LAP14}} & \multicolumn{2}{c|}{\textbf{REST14}} & \multicolumn{2}{c|}{\textbf{REST15}}  & \multicolumn{2}{c|}{\textbf{REST16}}  & \multicolumn{2}{c|}{\textbf{MAMS}} & \textbf{Training} \\ \\
        \addlinespace[0.05cm]
        \cline{2-11}
        \addlinespace[0.05cm]
        & ACC & F1 & ACC & F1 & ACC & F1 & ACC & F1 & ACC & F1 
        & \textbf{Time} \\
        \midrule
        Five-shots  
        & 60.56 & 60.84 
        & 65.29 & 64.52 
        & 62.64 & 54.60 
        & 69.24 & 60.15 
        & 65.31 & 62.52
        & 0.0 hours
        \\
        
        Full Tuning 
        & 82.76 & 78.59 
        & 85.71 & 76.90 
        & 86.06 & 73.92 
        & 91.24 & 82.49 
        & 81.32 & 81.70
        & 12.7 hours\\
        
        \midrule
        Strategy 1 &
\textbf{83.33} & \textbf{79.95} & \textbf{87.98} & \textbf{82.91} & \textbf{87.83} & \textbf{75.52} & \textbf{93.27} & \textbf{84.27} & \textbf{85.93} & \textbf{86.37}
        & 3.0 hours
        \\

        Strategy 2 
        & 83.27 & 80.11 
        & 87.68 & 82.86 
        & 87.47 & 75.56 
        & 93.21 & 84.45 
        & 86.20 & 85.73 
        & 2.3 hours
        \\
        
        \bottomrule
    \end{tabular}
}
\end{center}
\vspace{-0.3cm}
\caption{
\textcolor{black}{
The experiment results (i.e., accuracy and F1 scores) of models with different pluginning strategies on the test set of all datasets for ABSA.
We also report the training time (in hours) for different settings.
}
}
\label{tab: train}
\vskip -1em
\end{table*}

\textcolor{black}{
We further compare our best-performing model, i.e., LLaMA-2 with the plugins to model all different types of knowledge (LLaMA-2+P(C)+P(D)+P(S)), with previous studies, which also include a naive retrieval-augmented generation (RAG) approach that uses the retrieved syntactic knowledge and the input to prompt the LLM to predict the sentiment.
Table \ref{tab: sota} presents the results,
where our approach outperforms most existing studies and achieves state-of-the-art performance, demonstrating its effectiveness.
}

\subsection{The Results on Joint ABSA}

\textcolor{black}{
To explore the generalization capability of our approach to joint ABSA, where the model needs to predict the aspect term and the sentiment polarities at the same time, we run experiments on SemEval-2015.
We run the vanilla Qwen-2.5 baseline and our approach with three plugins\footnote{Since the aspect term is not given in this setting, we perform the following adaptation for obtaining different types of knowledge. For constituent syntax, we consider all phrases whose length is less than 10 words rather than focusing on the phrases that contain the aspect term; for word dependencies, we associate each word with the dependency relation type between that word and its head word and obtain the keys and values accordingly; for CCG supertags, we consider all words and their CCG supertags when building the keys and values without using a window.}.
The results of the baseline and our approach, as well as the performance of representative previous studies\footnote{\textcolor{black}{We use different baselines in Table \ref{tab: sota} and Table \ref{tab: sota-joint} because the tasks evaluated are different, where one approach works for one task may not able to be directly applied to the other.}}, are presented in Table \ref{tab: sota-joint}.
The results show that our approach outperforms the baseline and existing studies, and thus further confirms its effectiveness.
}

\subsection{The Effect of Memory Size}

\begin{figure}[t]
  \centering
  \includegraphics[width=0.5\textwidth, trim=0 40 0 0]{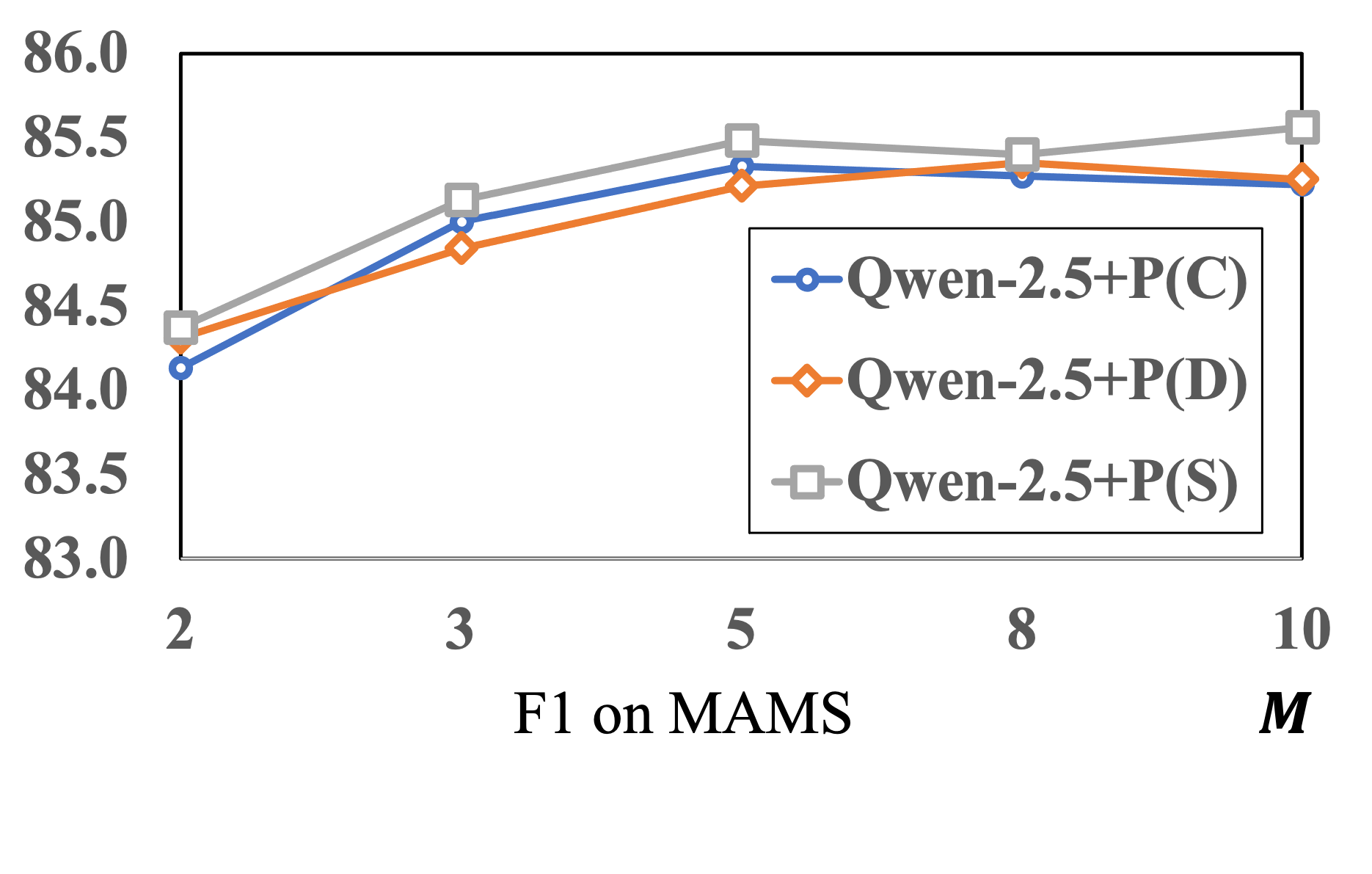}
  \caption{
  \textcolor{black}{
  The curves of F1 scores of ``+ P(C)'', ``+ P(D)'', and ``+ P(S)'' on MAMS with respect to the memory sizes in the plugin to leverage constituent syntax, word dependencies, and the CCG supertags.
  }
  }
  \vskip -1em
  \label{fig: memory size}
\end{figure}

\textcolor{black}{
To investigate the effect of the memory size, we run experiments using ``Qwen-2.5+P(C)'', ``Qwen-2.5+P(D)'', and ``Qwen-2.5+P(S)'' to test different memory sizes when particular knowledge is used.
We report the results (i.e., F1 scores) of the three models in Figure \ref{fig: memory size} when MAMS is used as the dataset to perform our investigation. 
For all cases, when the values are small (e.g., smaller than five), the model performance increases with the memory size enlarges.
This observation is intuitive since more context information is leveraged with larger memory size, which better instructs the LLM to perform ABSA.
When the memory sizes are large (e.g., greater than five), continually increasing the values does not bring further improvements.
The interpretation is the following.
For the syntactic information, increasing the memory size includes more low-frequent contextual features, which may introduce noise to the syntactic knowledge and thus make it hard for the plugin to distinguish the important knowledge from the unimportant ones.
}

\subsection{The Effect of Pluginning Strategy}
\label{sec: abaltion}

\textcolor{black}{
In the main experiments, we utilize the first approach (denoted as ``Strategy 1'') in Section \ref{sec: train} to tune the plugin and the hub when they are integrated with the LLM.
Given that the plugin is able to be trained without LLM, we run experiments with the second strategy (denoted as ``Strategy 2'') in Section \ref{sec: train}, where the plugin is firstly trained and then directly integrated with the LLM without further tuning.
For reference, we also report two other settings without using the plugin, where the first one (denoted as ``Five-shot'') utilizes the five-shot setting to prompt the LLM to perform ABSA, and the second one (denoted as ``Full Tuning'') fine-tunes all parameters in the vanilla LLM on the training data of ABSA.
The results of using Qwen-2.5 as the LLM are reported in Table \ref{tab: train}, where the average time (in hours) to train a model is also reported.
We observe that our approaches with the LLM plugin (i.e., Strategy 1 and 2) outperform the models with five-shot and full-tuning settings, which further confirms the effectiveness of our plugin design in leveraging different types of knowledge.
Meanwhile, our approaches require less training time compared with full tuning, which confirms that our approach is an effective solution for adapting LLMs to ABSA.
}

\subsection{Case Study}

\begin{figure}[t]
 \centering
 \includegraphics[width=0.5\textwidth, trim=0 20 0 0]{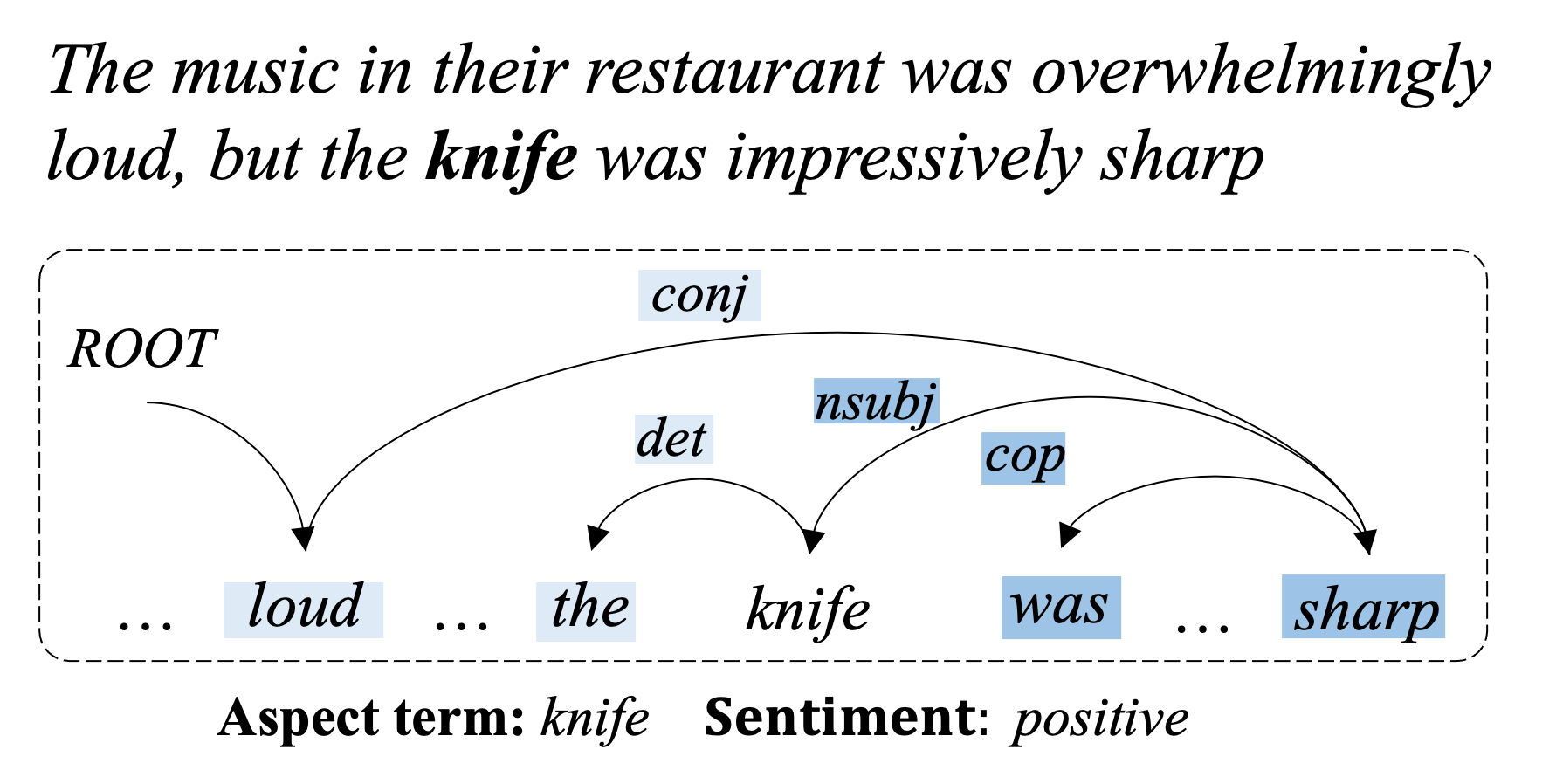}
  \caption{
  \textcolor{black}{An example sentence with the aspect term ``\textit{knife}'' highlighted in boldface. The background colors illustrate the weights assigned for different dependencies in the memory mechanism of the plugin, where deeper colors refer to higher weights.}
  }
  \vskip -1em
  \label{fig: case}
\end{figure}

\textcolor{black}{
To illustrate how our plugin enhances LLM for ABSA, we perform a case study with the word dependencies.
Figure \ref{fig: case} presents the example sentence with an aspect term ``\textit{knife}'', whose gold standard sentiment is \textit{positive} and our approach is able to correctly predict the sentiment label.
In this figure, we also present a part of the dependency tree that is relevant to the aspect term.
The background colors of the dependencies illustrate the weights for different word-word connections in the dependency plugin, where deeper colors refer to higher weights.
From the figure, we find that the memory mechanism is able to assign high weights to dependencies (e.g, the first-order dependency with ``\textit{sharp}'') that are important for predicting the sentiment of ``\textit{knife}'' and assign low weights to the unimportant ones (e.g., the second-order dependency with ``\textit{loud}'').
Therefore, based on the dependency information, the model is able to leverage the information for predicting the sentiment label correctly.
}

\section{Related Work}

Different from sentence-level sentiment analysis, ABSA is a more fine-grained and entity-level oriented task that aims to determine the sentiment polarities of given aspects in a sentence.
Recently, various attention-based neural networks have been proposed to capture the contextual information, especially the aspect term and its contexts \cite{xue2018aspect,li-etal-2018-transformation,mao2019aspect,xu-etal-2019-bert,jiang-etal-2019-challenge,xu-etal-2020-aspect,tang-etal-2020-dependency,wang-etal-2020-relational,qin-2022-complementary,tian-etal-2024-aspect-based}.
Besides advanced decoders, another mainstream trend is incorporating syntactic knowledge to establish the structural connection between the aspect and its corresponding opinion words and enhance text representation.
In addition, the effort devoted into combining graph neural networks (e.g., GCN, GAT) and syntactic information, e.g., dependency tree from off-the-self dependency parsers, have shown gratifying results in ABSA \cite{zhang2019aspect,wang-etal-2020-relational,zhang-qian-2020-convolution,liang-etal-2021-iterative-multi,zhang-etal-2022-ssegcn,zhang-etal-2023-span,qin-etal-2021-improving-federated}.
\textcolor{black}{
Moreover, there are recent studies that leverage LLM to improve ABSA \cite{simmering2023large,luo2024panosent,negi2024hybrid,ding2024boosting,zhu2024zzu}, where further enhancements such a data augmentation \cite{li2024iterative} and similar instance retrieval \cite{zheng2024instruction} are applied.
}
\textcolor{black}{
Compared with previous studies, our approach utilizes the LLM plugin to leverage different types of knowledge, such as the constituent syntax, word dependencies, and CCG supertags of the sentence, to improve ABSA.
In addition, a memory mechanism is proposed to encode different knowledge instances, which allows the model to identify the important ones and leverage them accordingly.
}

\section{Conclusion}

In this paper, we propose an LLM-based approach for ABSA, where a memory-based plugin is proposed to enhance it with requiring limited computation and resources, also compatible with different types of knowledge.
\textcolor{black}{
Specifically,
we evaluate our approach with using constituent syntax, word dependencies, and CCG supertags,
and experiment results on benchmark datasets demonstrate its effectiveness, which outperforms strong baselines and previous studies and achieves state-of-the-art results on all datasets.
}

\bibliography{custom,llm,absa-llm}

\end{document}